# Identifying roadway departure crash patterns on rural two-lane highways under different lighting conditions: association knowledge using data mining approach

**Ahmed Hossain**
*(Corresponding Author)*
Ph.D. Student
Department of Civil Engineering
University of Louisiana at Lafayette, Lafayette, Louisiana, 70504
Email: ahmed.hossain1@louisiana.edu
ORCID ID: 0000-0003-1566-3993

**Xiaoduan Sun, Ph.D., P.E.**
Professor
Department of Civil Engineering
University of Louisiana at Lafayette, Lafayette, Louisiana, 70504
Email: xiaoduan.sun@louisiana.edu
ORCID ID: 0000-0001-7282-1340

**Shahrin Islam**
Graduate Student
Department of Civil Engineering
Bangladesh University of Engineering and Technology, Bangladesh, 1000
Email: shahrin.ce@diu.edu.bd
ORCID ID: 0000-0001-9942-5794

**Shah Alam**
Lecturer
Department of Civil Engineering
Rajshahi Science & Technology University, Natore, Bangladesh, 6400
Email : shahalamruet4@gmail.com
ORCID ID: 0000-0001-8766-1503

**Md Mahmud Hossain**
Ph.D. Student
Department of Civil and Environmental Engineering, Auburn University
218A Ramsay Hall, Auburn, AL 36849-5337
Email: mahmud@auburn.edu
ORCID ID: 0000-0001-8766-1503

**Abstract**

*Introduction:* More than half of all fatalities on the U.S. highways occur due to roadway departure (RwD) each year. Previous research has explored various risk factors that contribute to RwD crashes, however, a comprehensive investigation considering the effect of lighting conditions has been insufficiently addressed. *Data:* Using the Louisiana Department of Transportation and Development crash database, fatal and injury RwD crashes occurring on rural two-lane (R2L) highways between 2008-2017 were analyzed based on daylight and dark (with and without streetlight). Method: This research employed a safe system approach to explore meaningful complex interactions among multidimensional crash risk factors. To accomplish this, an unsupervised data mining algorithm association rules mining (ARM) was utilized. *Results and conclusions:* Based on the generated rules, the findings reveal several interesting crash patterns in the daylight, dark-with-streetlight, and dark-no-streetlight, emphasizing the importance of investigating RwD crash patterns depending on the lighting conditions. In daylight condition, fatal RwD crashes are associated with the cloudy weather condition, distracted drivers, standing water on the roadway, no seat belt use, and construction zones. In dark lighting condition (with and without streetlight), the majority of the RwD crashes are associated with alcohol/drug involvement, young drivers (15-24 years), driver condition (e.g., inattentive, distracted, illness/fatigued/asleep) and colliding with animal (s). *Practical Applications:* The findings also reveal how certain driver behavior patterns are connected to RwD crashes, such as a strong association between alcohol/drug intoxication and no seat belt usage in the dark-no-streetlight condition. Based on the identified crash patterns and behavioral characteristics under different lighting conditions, the findings could aid researchers and safety specialists in developing the most effective RwD crash mitigation strategies.

# 1. Introduction

Despite the fact that the number of miles driven reduces extensively at night compared to the day, more than half of all the traffic fatalities occur in dark lighting conditions (Plainis, Murray, & Pallikaris, 2006). As a result, there is a growing interest in understanding the relationship between lighting conditions and traffic crashes on roadways (Jägerbrand & Sjöbergh, 2016). Since Roadway Departure (RwD) is the most common type of traffic collision, studying its relationship with various illumination conditions can lead to critical knowledge discoveries and improved roadway safety. According to the Federal Highway Administration (FHWA), a RwD crash occurs when a vehicle crosses an edge line or a centerline or otherwise leaves the travel path (McGee Sr, 2018). According to the Fatality Analysis Reporting System (FARS), from 2016 through 2018, the RwD crash caused an average of 19,158 fatalities, accounting for approximately half of all the traffic fatalities in the United States. RwD crashes are also a major concern in Louisiana State, particularly on rural two-lane (R2L) highways. According to the Louisiana Department of Transportation and Development (LADOTD) crash database, a total of 83,843 crashes on R2L highways have been reported in Louisiana State between 2008 and 2017, with RwD accounting for approximately 65% of them. The Louisiana Strategic Highway Safety Plan (SHSP) has identified RwD crashes as one of the five focus areas. To reach the 'Destination Zero Deaths' goal set by Louisiana SHSP, the target is to minimize the number of people killed or seriously injured due to RwD crashes by half within the 2030 (LADOTD, 2018). Therefore, identifying the crash risk factors of RwD crashes is critical to achieving the desired goal.

Lighting condition has been recognized as an important factor associated with RwD crashes in the previous research (Hashemi & Archilla, 2017; Lord, Brewer, Fitzpatrick, Geedipally, & Peng, 2011; Rahman, Sun, Das, & Khanal, 2021; Wang & Knipling, 1994; Zhu, Dixon, Washington, & Jared, 2010). Most of the prior research revealed that single-vehicle run-off-road (SVROR), a type of RwD crash, was more prevalent in a location with dark lighting conditions than in locations with better lighting circumstances or daytime conditions. Additionally, the authors also reported that RwD crashes that occurred during daylight or dark-lighted conditions were less likely to result in fatalities. A review of RwD crash injury severity (K= Fatal, A = Severe, B = Moderate, C = Complaint, O = No injury) on R2L highways in Louisiana (2008-2017) according to different lighting conditions (shown in **Figure 1**) depicts that, fatalities in the dark-no-streetlight condition is more compared to the daylight (52.1% vs 40.4%). Additionally, the percentage of severe injury crashes is comparatively similar compared to daylight (44.5%) and dark-no-streetlight conditions (44.9%). Interestingly, RwD collisions in the dark with streetlights account for a relatively small fraction of crashes through all severity levels. As a result, it is critical to look at the RwD crashes under various lighting settings, as this can assist in a better understanding of the crash patterns. Note that, the percentages in the following figure do not add up to 100% (row-wise) because some of the crashes occurred in dusk/dawn or 'unknown' lighting conditions.

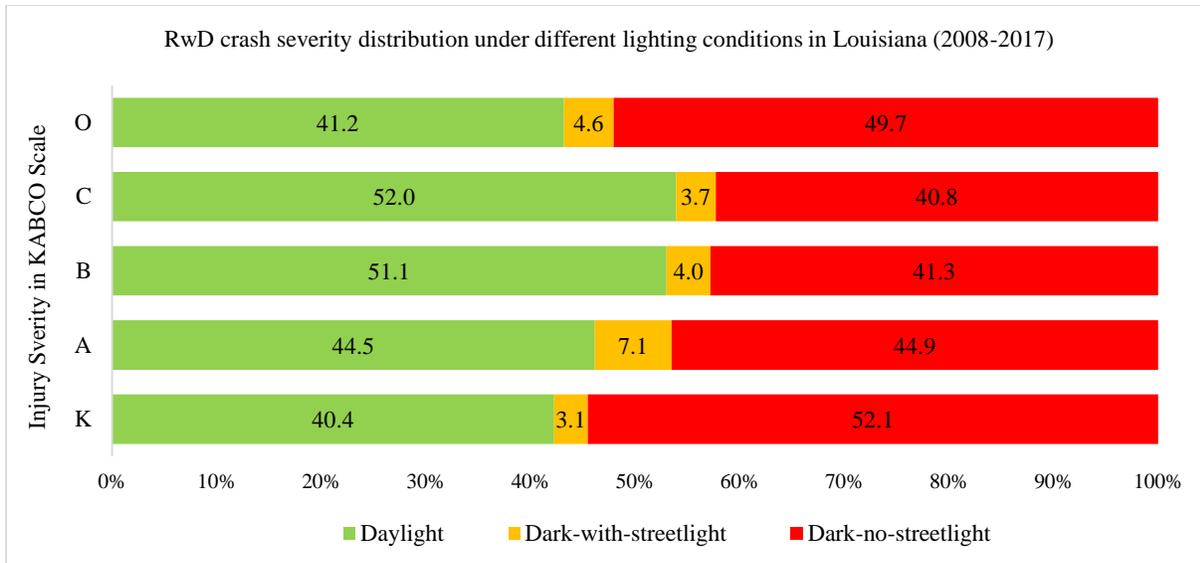

Figure 1. Crash injury severity of RwD crashes in Louisiana (2008-2017)

Each crash is caused by a chain of critical events. RwD collisions on R2L roadways would most likely be caused by a mix of attributes rather than a single associated attribute. Due to the dynamic and complex nature of a RwD crash, different components such as human and vehicle-related factors, roadway geometric features, and environmental settings actively interact with each other to cause a crash (Appiah & Zhao, 2020). Moreover, the pattern and outcome of RwD crashes vary significantly between the lighted and dark conditions (Al-Bdairi, Hernandez, & Anderson, 2018). Gaining this knowledge of interaction and association of risk factors separately under daylight and dark condition can be a major step towards improving safety on R2L highways. This study focuses on investigating the patterns in RwD crashes under different lighting conditions, intending to present the collective associations of crash contributing factors that have not been previously explored. To account for the categorical data in the crash dataset, this study utilized Association Rules Mining (ARM), a data mining approach to capture the simultaneous presence of crash contributing factors.

## 2. Literature Review

Well-studied RwD crash influencing factors are linked with human behavioral characteristics, roadway geometry, traffic attributes, environmental circumstances, and pavement surface condition (Al-Bdairi & Hernandez, 2017; Al-Bdairi et al., 2018; Eustace, Almutairi, & Hovey, 2016; Turochy & Ozelim, 2016). More than two decades earlier, Wang and Knipling analyzed single-vehicle roadway departure (SVRD) crashes using the FARS database (Wang & Knipling, 1994). They discovered that the major share of SVRD crashes occurred during the day, with fatal crashes occurring most frequently between 12 a.m. and 6 a.m. Lord et al. utilized six years (2003-2008) of crash data from rural two-lane highways in Texas to evaluate the factors that contribute to RwD crash (Lord et al., 2011). The study identified a higher proportion of nighttime run-off-road (ROR) crashes compared to the daytime including fatal crashes, crashes on curves, crashes caused due to drivers attempting to avoid colliding with an animal(s), and related to impaired/fatigue/asleep drivers. Another research conducted in California investigated driver injury severity involved in single-vehicle crashes (Kim, Ulfarsson, Kim, & Shankar, 2013). They identified lighting conditions as a critical factor that affects crash risk and darkness (with or without streetlight) increases the risk of fatalities and serious injuries while involved in single-vehicle crashes. It is worth noting that the majority of RwD collisions include single-vehicle (SV), or single-vehicle run-off-road (SVROR) crashes (X. Sun & Rahman, 2021).

Excessive speed has been identified as a key contributor to RwD collisions (Council, Reurings, Srinivasan, Masten, & Carter, 2010). According to a recent FHWA report, the period from 6 p.m. to 5 a.m. was identified as the most dangerous, with approximately 60% of RwD crashes occurring during this time (Neuner et al., 2016). A study by Massie and Campbell found that the probability of being involved in a single-vehicle accident at night is 5.3 times higher for male drivers and 3.4 times higher for female drivers than it was during the day (Massie, 1993). The higher the blood alcohol concentration (BAC) of a driver, the greater the probability that the crash involves a single-vehicle (National Highway Traffic Safety Administration, 2002). Relative to non-drinking drivers, drivers in all age and gender groups with BAC of 0.08 to 0.099% had at least 11 times higher risk of fatalities involved in a single-vehicle crash at night during the weekend (Hingson & Winter, 2003).

The majority of the aforementioned previous studies utilized 'lighting conditions' as a single indicator variable and did not distinguish between the crashes occurring in daylight and nighttime. Such a strategy has limitations because variable interaction affects RwD crash outcomes in a dynamic and complicated manner depending on the illumination circumstances (Pahukula, Hernandez, & Unnikrishnan, 2015). For example, the driver's driving behavior (e.g., speed, acceleration, lane-keeping), as well as other human factors (e.g., fatigue, drinking-and-driving), is significantly different considering the daylight and dark condition (Gibbons, Milton, & van Schalkwyk, 2016; K. Zhang & Hassan, 2019). The surrounding associated environmental settings also tend to change as the temperature drops at night, and roads are more prone to turn slippery at night than during the day. Therefore, a simple indicator variable-based strategy would fail to address the intricacies of variable interactions based on different lighting conditions. To account for these fluctuations, different models (i.e., daylight, dark) must be developed so that the precise effect of human, vehicle, roadway and environmental factors affecting RwD crashes under different lighting conditions be able to be determined and reveal more intuitive crash patterns.

To explore the interconnections between the risk factors and RwD crashes, the majority of the previous literature applied either the frequency distribution analysis (Kusano & Gabler, 2013) or other parametric models such as the logistic regression (Spainhour & Mishra, 2008), mixed logit model (Islam & Pande, 2020), multinomial logistic model (Peng, Geedipally, & Lord, 2012), ordered probit model (M. Yu, Ma, & Shen, 2021), and negative binomial regression model (Lord et al., 2011). However, these parametric models have been chastised for their predetermined presumption (e.g., dependent variables must be mutually exclusive), which may not always be true (S. Yu, Jia, & Sun, 2019). Association Rules Mining (ARM) method can be a viable alternative as no variable is selected dependent or independent and it can reveal significant insights from a complex multidimensional crash database (M. M. Hossain, Sun, Mitran, &

Rahman, 2021). In light of this, the study employed ARM, a non-parametric unsupervised data mining approach that allows discovering the relationships among crash risk factors without making any prior assumptions about the variables.

Based on the existing review of literature, investigating the pattern of risk factors in RwD crashes that occurred on R2L highways has been overlooked considering the effect of lighting conditions. This study intends to fill the gap in the literature by addressing the effect of lighting conditions in RwD crashes on R2L highways. Unlike most of the prior studies, which looked at lighting conditions as a single contributing factor, this study divides lighting conditions into three groups (daylight, dark-with-streetlight, dark-no-streetlight) to determine the underlying pattern of RwD crashes. Lastly, this study can offer safety officials comprehensive knowledge about the impact of lighting conditions in R2L RwD crashes and assist them to select appropriate countermeasures.

## *2.1 Primary Variable Selection*

The variable selection procedure comprises a comprehensive assessment of relevant literature and the identification of important contributing variables that impact RwD crashes. **Table 1** below highlights a list of research, their study area, data, methodologies, and crash variables.

Table 1. List of literature on RwD crash analysis

| Study and location | Data and Methods | Variable Utilized |
|---|---|---|
| (Rahman et al., 2021); Louisiana | 2005-2017; Logit model and association rules mining | Time of the day, day of the week, lighting condition, surface condition, weather, roadway condition, AADT, lane width, shoulder width, curve radius, vertical alignment, posted speed limit, driver age, driver gender, driver license state, driver distraction, alcohol, drug, passenger presence, vehicle type, vehicle year. |
| (Hashemi & Archilla, 2017); Oahu, Hawaii | 2008-2011; Classification and Regression Tree (CART) | Area type, road owner, weather, lighting conditions, land use, road classification, horizontal alignment, vertical alignment, pavement surface, surface wetness, number of lanes, speed limit. |
| (Appiah & Zhao, 2020); Virginia | 2014-2018: Principal Component Analysis (PCA), Multinomial logit regression, negative binomial (NB), and zero-inflated Poisson (ZIP) regression models. | Segment length, AADT, roadway geometry, shoulder condition, surface condition, speed limit, functional class, traffic control, weather condition, median width. |
| (Islam & Pande, 2020); Minnesota | 2010-2014; Random parameter logit model | Vehicle type, AADT, driver gender, driver age, speeding, surface condition, weather condition, horizontal alignment. |
| (Lord et al., 2011); Texas | 2003-2008; Poisson regression model | Weather condition, lighting condition, day of the week, time of the day, presence of shoulder, |

| | | shoulder type, lane width, shoulder width, driveway density, curve density, ADT. |
|---|---|---|
| (Peng et al., 2012); Texas | 2003-2008; Negative binomial regression, multinomial logit regression | Length of section, ADT, shoulder width, side slope rating, driveway density, lateral clearance, crash type |

Some of the variables (e.g., driveway density, road owner, side slope rating) utilized in previous research were not available in the Louisiana crash database. Based on engineering judgment, availability, and the review of previous literature, the following variables were selected for this study –
- Crash characteristics (injury severity, manner of collision)
- Roadway characteristics (AADT, road condition, lane width, shoulder width, curve radius, vertical alignment, speed limit)
- Driver and vehicle characteristics (driver age, driver gender, driver protection system, driver condition, passenger presence, vehicle type).
- Environmental characteristics (lighting condition, surface condition, weather condition)
- Temporal factors (day of the week).

## 3. Methodology

### 3.1 Safe System Approach

The fundamental idea of the safe system approach is to manage human, vehicle, roadway, and environmental factors and the interaction among these components to minimize the likelihood of fatalities and serious injuries (Langford & Oxley, 2006). A similar idea is implemented in this research to reveal the interaction of contributing factors leading to RwD crashes on R2L highways. To address the effect of lighting conditions, the analysis is undertaken considering three different lighting conditions (daylight, dark-with-streetlight, dark-no-streetlight). Therefore, this research will investigate how the association of crash contributing factors varies according to three different lighting conditions and identify appropriate countermeasures based on the obtained crash patterns. To accomplish this, an unsupervised data mining method Association Rules Mining (ARM) is utilized.

### 3.2 Association Rules Mining

Association Rules Mining (ARM), a rule-based machine learning algorithm, can identify frequent itemset that occurs together in an event. Here, 'itemset' refers to the collection of variable categories and an 'event' is analogous to a RwD crash in this case. The 'Apriori' algorithm, the most popular framework for applying ARM, was used by the research team (Agrawal, Imieliński, & Swami, 1993). At first, ARM was utilized for market basket analysis to determine the pattern in which customers made purchases (Ünvan, 2021). It has become a more common approach to recognizing crash patterns in the highway safety research (A. Hossain, Sun,

Thapa, & Codjoe, 2022; M. M. Hossain, Rahman, Sun, & Mitran, 2021; M. M. Hossain, Zhou, Das, Sun, & Hossain, 2022; Hsu & Chang, 2020; Rahman et al., 2021).

In this current study, an 'item' is defined as one component of an itemset, which symbolizes a reported RwD crash. Assume I = $\{i_1, i_2, \ldots, i_n\}$ as a finite set of items and T = $\{t_1, t_2, \ldots, t_m\}$ be a crash database of transactions that are a subset of I. An association rule can be written as Antecedent (X) → Consequent (Y). Here, antecedent (X) is a set of one or more items, and consequent is a single item, both of which belong to crash database T. Note that, a consequent is an item that is found in combination with the antecedent. For example, {driver_age=>64} → {lighting condition = daylight} is an association rule, where {driver_age=>64} is the antecedent (X) which appears in the L.H.S (Left Hand Side), and {lighting condition = daylight} is the Consequent (Y) which appears in the R.H.S. (Right Hand Side) of the rule. It is important to mention here that, antecedent and consequent are disjoint, that is, they have no items in common (X ∩ Y = Ø).

In ARM, support and confidence are two critical parameters for selecting important rules from a large number of possible rules (Hong, Tamakloe, & Park, 2020). The support (S) indicates how frequently the antecedent (X) and consequent (Y) of a given rule occur together in the database, while the confidence (C) evaluates the rule strength by estimating the probability P(X|Y). The equation of support and confidence is given below.

$$support\ (X \to Y) = P\ (X \cap Y) = \frac{|X \cup Y|}{|T|} \quad (1)$$

$$confidence\ (X \to Y) = \frac{support\ (X \to Y)}{support\ (X)} = \frac{P(X \cap Y)}{P(X)} \quad (2)$$

ARM algorithms may generate a large number of rules that satisfy the predefined support and confidence levels, depending on the dataset being investigated (Weng, Zhu, Yan, & Liu, 2016). In this viewpoint, another measure known as the 'Lift' was proposed to address the constraints. The equation for lift is shown in the following equation.

$$lift\ (X \to Y) = \frac{confidence\ (X \to Y)}{support\ (X)} = \frac{P(X \cap Y)}{P(X)\ P(Y)} \quad (3)$$

The parameter lift (L) represents the relationship between the antecedent-consequent co-occurrence frequency and the predicted frequency (Kotsiantis & Kanellopoulos, 2006). There could be three possible scenarios concerning the Lift value (<1, =1, >1). With a lift equal to one, there is no relationship between itemset X and Y, whereas itemset X is exclusive to item set Y with a lift value less than one. The association rules are recognized as significant only if the lift is greater than one. The analysis was carried out with the help of the open-source program R version 4.0.1 and the R package 'arules' (Hahsler, Chelluboina, Hornik, & Buchta, 2011). Another R package 'arulesViz' was utilized for interactive visualization of association rules (Hahsler, 2017).

*3.3 Explanation of Association rules*

A real example from the crash dataset used in this study is provided below to demonstrate association rules. The database used in this study has a total of 7,568 rows. Since association rules can be written as Antecedent (X) → Consequent (Y), let us consider the association rule in a similar format: {driver_age=>64} → {lighting condition = daylight}. The calculation for support, confidence, and lift is summarized in the following **Table 2**.

Table 2. Calculations of association rules parameters

| Total rows in the entire database | 7,568 |
|---|---|
| Lighting condition = daylight appeared | 3,851 times |
| Within daylight lighting conditions, driver age group '>64 years' appeared | 282 times |
| Support (S) of the association rule | $= \frac{282}{7568} = 3.73\%$ |
| Driver age group '>64 years' appeared in the entire database | 385 times |
| Confidence (C) of the association rule | $= \frac{282}{385} = 73.24\%$ |
| Lift (L) of the association rule | $= (282 \div 7568)/[(385 \div 7568) \times (3851 \div 7568)] = 1.44$ |

The explanation of the above-mentioned association rule is: '>64 years' aged drivers account for 3.73% of the RwD crashes that occur in the daylight condition; out of all the '>64 years' aged drivers involved in RwD crashes occurred on roadways, 73.24% took place in the daylight condition; the proportion of '>64 years' aged drivers involvement in crashes occurring on roadways during the daylight is 1.44 times the proportion of all '>64 years' aged drivers involved crashes in the entire dataset.

*3.4 Exploratory data analysis (EDA)*

The Louisiana Department of Transportation and Development (LADOTD) crash database was utilized to pull all the crashes reported by police between 2008 and 2017. The primary database was created by merging four data tables (crash table, vehicle table, highway table, and DOTD table) with the help of using a 'crash number' as a matching criterion. The crash database utilizes a variable 'HIGHWAY_CLASS' which was set to '1' (Rural 2-lane) to select only the crashes that occurred on R2L highways. The crash database uses an 'ABDEFYZ' scale (A = Daylight, B = Dark-no-streetlight, C = Dark-continuous streetlight, D = Streetlight at intersection only, E = Dusk, F = Dawn, Y = Unknown, Z = Other) to designate the crash lighting conditions. The first category 'A', i.e., 'daylight' is simply defined by when natural light is present and is not otherwise considered dawn or dusk. This study excluded the 'dawn' and 'dusk' category because it is considered that reporting at these times of day is objective and more likely to be misattributed (Terry, Brimley, Gibbons, & Carlson, 2016). The other three categories 'B', 'C', and 'D' encompass driving scenarios at dark condition (with or without streetlight). The crash database utilizes an 'ABCDE' scale (A = Fatal, B = Severe, C = Moderate, D = Complaint, E = No injury) to categorize injury severity in a crash. The research team chose only fatal, severe, and moderate injury crashes to focus on only the evident injury crashes. It is worth mentioning that injury categories 'D' and 'E' do not exhibit any physical evidence of injury (National Safety Council, 2010). The final database was prepared using a three-step filtering process (filter 1: RwD = yes, filter 2: injury severity = fatal/severe/moderate, filter 3: lighting condition = daylight, dark). The final database contains 7,568 unique RwD crashes (fatal/severe/moderate) that occurred in three different lighting conditions i.e., daylight (3851, 50.89%) and dark-with-streetlight (323, 4.26%), and dark-no-streetlight (3394, 44.85%). Data integration and analysis flowchart is shown in the **Figure 2**.

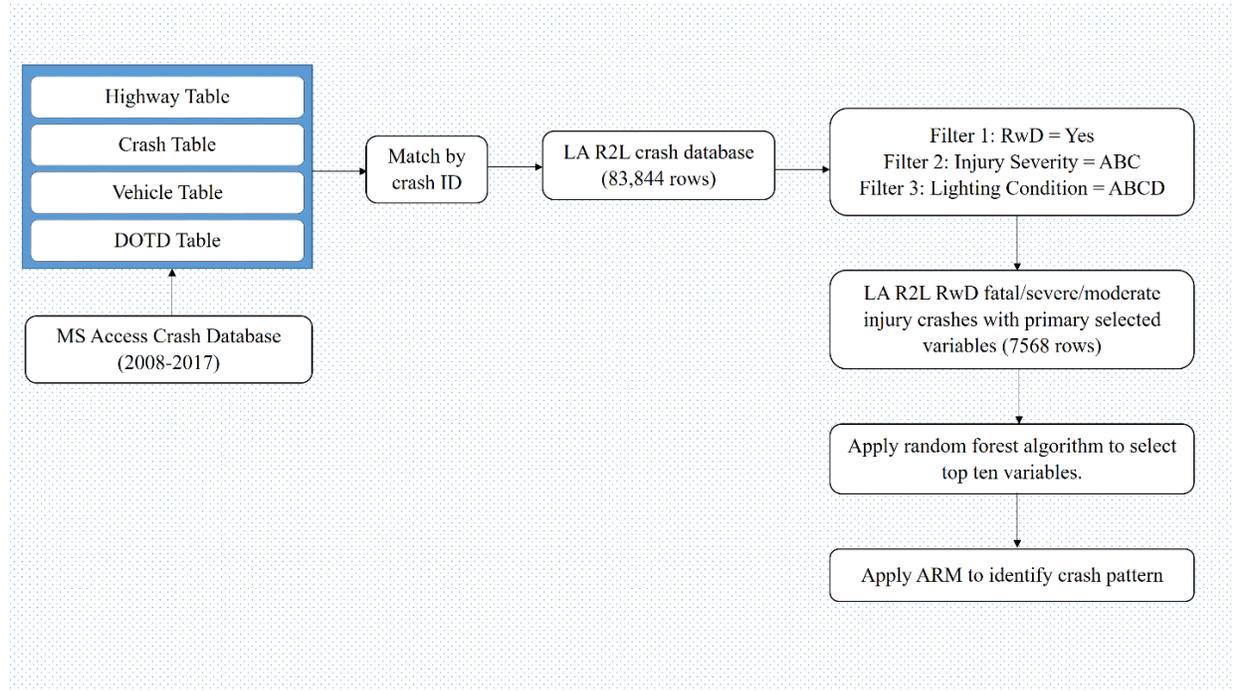

Figure 2. Data preparation and analysis flowchart

To analyze the association of contributing factors causing RwD crashes according to daylight and dark (with and without streetlight), nineteen crash variables were primarily selected based on earlier RwD-related safety research. The following **Table 3** provides descriptive statistics of each crash variable category according to three different lighting conditions. Note that percentages in the following table may not add up to 100% due to rounding errors.

Table 3. Overview of RwD crashes by lighting conditions

| Type | Variable Label | Variable Categories | Daylight, N (%) | Dark, N (%) | |
|---|---|---|---|---|---|
| | | | | With streetlight | No streetlight |
| Crash Characteristics | injury_severity | fatal | 485 (12.59%) | 37 (11.46%) | 626 (18.44%) |
| | | severe | 246 (6.39%) | 39 (12.07%) | 248 (7.31%) |
| | | moderate | 3120 (81.02%) | 247 (76.47%) | 2520 (74.25%) |
| | manner_of_collision | single_vehicle | 3470 (90.11%) | 288 (89.16%) | 3234 (95.29%) |
| | | head_on | 107 (2.78%) | 4 (1.24%) | 33 (0.97%) |
| | | rear_end | 24 (0.62%) | 3 (0.93%) | 6 (0.18%) |
| | | right_angle | 59 (1.53%) | 1 (0.31%) | 21 (0.62%) |
| | | right_left_turn | 11 (0.29%) | 1 (0.31%) | 5 (0.15%) |
| | | sideswipe | 68 (1.77%) | 4 (1.24%) | 22 (0.65%) |
| | | others | 112 (2.91%) | 22 (6.81%) | 73 (2.15%) |
| Environmental and temporal factors | surface_condition | dry | 3252 (84.45%) | 268 (82.97%) | 2884 (84.97%) |
| | | non_dry | 599 (15.55%) | 55 (17.03%) | 502 (14.79%) |
| | | unknown | 0 (0%) | 0 (0%) | 8 (0.24%) |
| | weather_condition | clear | 2799 (72.68%) | 240 (74.30%) | 2544 (74.96%) |
| | | cloudy | 620 (16.10%) | 42 (13.00%) | 486 (14.32%) |
| | | rain | 388 (10.08%) | 29 (8.98%) | 268 (7.90%) |
| | | snow_sleet_hail | 18 (0.47%) | 2 (0.62%) | 7 (0.21%) |
| | | others | 26 (0.68%) | 10 (3.10%) | 89 (2.62%) |
| | day_of_week | weekday | 2616 (67.93%) | 178 (55.11%) | 1941 (57.19%) |
| | | weekend | 1235 (32.07%) | 145 (44.89%) | 1453 (42.81%) |
| Roadway Characteristics | aadt | <400 | 196 (5.09%) | 8 (2.48%) | 164 (4.83%) |
| | | 401-1000 | 598 (15.53%) | 31 (9.60%) | 525 (15.47%) |
| | | 1001-5000 | 2448 (63.57%) | 192 (59.44%) | 2125 (62.61%) |
| | | >5000 | 609 (15.81%) | 92 (28.48%) | 580 (17.09%) |
| | road_condition | no_abnormalities | 3556 (92.34%) | 298 (92.26%) | 3049 (89.84%) |
| | | shoulder_abnormality | 30 (0.78%) | 1 (0.31%) | 13 (0.38%) |
| | | standing_water | 52 (1.35%) | 2 (0.62%) | 127 (3.74%) |
| | | animal | 12 (0.31%) | 0 (0%) | 34 (1.00%) |
| | | construction | 21 (0.55%) | 3 (0.93%) | 17 (0.50%) |
| | | previous_crash | 1 (0.03%) | 0 (0%) | 2 (0.06%) |
| | | others | 129 (3.35%) | 7 (2.17%) | 105 (3.09%) |
| | | unknown | 50 (1.30%) | 12 (3.72%) | 47 (1.38%) |
| | lane_width (unit = ft.) | <11 | 935 (24.28%) | 75 (23.22%) | 786 (23.16%) |
| | | 11<=lw<12 | 1508 (39.16%) | 113 (34.98%) | 1304 (38.42%) |
| | | >=12 | 1121 (29.11%) | 105 (32.51%) | 1008 (29.70%) |
| | | unknown | 287 (7.45%) | 30 (9.29%) | 296 (8.72%) |
| | shoulder_width (unit = ft.) | <=2ft | 764 (19.84%) | 46 (14.24%) | 652 (19.21%) |
| | | 2ft<x<=4ft | 1124 (29.19%) | 107 (33.13%) | 989 (29.14%) |
| | | 4ft<x<=6ft | 580 (15.06%) | 45 (13.93%) | 510 (15.03%) |
| | | >6ft | 1154 (29.97%) | 96 (29.72%) | 998 (29.40%) |
| | | unknown | 229 (5.95%) | 29 (8.98%) | 245 (7.22%) |
| | curve_radius (unit = ft.) | tangent | 2172 (56.40%) | 186 (57.59%) | 1945 (57.31%) |
| | | <=500 | 137 (3.56%) | 17 (5.26%) | 110 (3.24%) |
| | | 501_to_1000 | 376 (9.76%) | 33 (10.22%) | 337 (9.93%) |
| | | 1001_to_2500 | 663 (17.22%) | 30 (9.29%) | 594 (17.50%) |
| | | 2501_to_5000 | 327 (8.49%) | 28 (8.67%) | 249 (7.34%) |
| | | 5001_to_10000 | 176 (4.57%) | 29 (8.98%) | 159 (4.68%) |
| | vertical_alignment | level | 3088 (80.19%) | 273 (84.52%) | 2847 (83.88%) |
| | | level_elevated | 172 (4.47%) | 19 (5.88%) | 102 (3.01%) |
| | | on_grade | 420 (10.91%) | 12 (3.72%) | 319 (9.40%) |
| | | dip_hump | 7 (0.18%) | 3 (0.93%) | 5 (0.15%) |
| | | hillcrest | 106 (2.75%) | 4 (1.24%) | 67 (1.97%) |
| | | other | 5 (0.13%) | 0 (0%) | 3 (0.09%) |
| | | unknown | 53 (1.38%) | 12 (3.72%) | 51 (1.50%) |
| | speed_limit (unit = mph) | <=35 | 115 (2.99%) | 51 (15.79%) | 59 (1.74%) |
| | | 40<= x <=55 | 3623 (94.08%) | 258 (79.88%) | 3268 (96.29%) |
| | | >55 | 25 (0.65%) | 1 (0.31%) | 9 (0.27%) |
| | | unknown | 88 (2.29%) | 13 (4.02%) | 58 (1.71%) |
| Driver & vehicle characteristics | driver_age | 15-24 | 1098 (28.51%) | 102 (31.58%) | 1177 (34.68%) |
| | | 25-34 | 787 (20.44%) | 85 (26.32%) | 886 (26.10%) |
| | | 35-44 | 571 (14.83%) | 53 (16.41%) | 553 (16.29%) |
| | | 45-54 | 633 (16.44%) | 42 13.00%) | 442 (13.02%) |
| | | 55-64 | 435 (11.30%) | 26 (8.05%) | 192 (5.66%) |
| | | >64 | 282 (7.32%) | 6 (1.86%) | 97 (2.86%) |
| | | unknown | 45 (1.17%) | 9 (2.79%) | 47 (1.38%) |
| | driver_gender | male | 2508 (65.13%) | 249 (77.09%) | 2517 (74.16%) |
| | | female | 1316 (34.17%) | 69 (21.36%) | 827 (24.37%) |
| | | unknown | 27 (0.70%) | 5 (1.55%) | 50 (1.47%) |
| | driver_protection_system | properly_used | 2507 (65.10%) | 169 (52.32%) | 1718 (50.62%) |
| | | improperly_used | 31 (0.80%) | 3 (0.93%) | 13 (0.38%) |
| | | none_used | 1089 (28.28%) | 99 (30.65%) | 1402 (41.31%) |
| | | unknown | 224 (5.82%) | 52 (16.10%) | 261 (7.69%) |
| | driver_condition | normal | 944 (24.51%) | 58 (17.96%) | 930 (27.40%) |
| | | inattentive | 869 (22.57%) | 29 (8.98%) | 526 (15.50%) |
| | | distracted | 125 (3.25%0 | 4 (1.24%) | 103 (3.03%) |
| | | ill_fatigued_asleep | 360 (9.355) | 8 (2.48%) | 256 (7.54%) |
| | | alcohol | 156 (4.05%) | 44 (13.62%) | 451 (13.29%) |
| | | drug | 60 (1.56%) | 3 (0.93%) | 49 (1.44%) |
| | | other | 1337 (34.72%) | 177 (54.80%) | 1079 (31.79%) |
| | passenger_present | yes | 1088 (28.25%) | 101 (31.27%) | 967 (28.49%) |
| | | no | 2756 (71.57%) | 220 (68.11%) | 2421 (71.33%) |
| | | unknown | 7 (0.18%) | 2 (0.62%) | 6 (0.18%) |
| | vehicle_type | car_van_SUV | 1942 (50.43%) | 169 (52.32%) | 1839 (54.18%) |
| | | light_truck | 1155 (29.99%) | 128 (39.63%) | 1265 (37.27%) |
| | | truck | 233 (6.05%) | 3 (0.93%) | 57 (1.68%) |
| | | bus | 8 (0.21%) | 0(0%) | 1 (0.03%) |
| | | others | 513 (13.32%) | 23 (7.12%) | 232 (6.84%) |

The table of descriptive statistics revealed several important crash characteristics comparing daylight and dark (with and without streetlight). At dark no streetlight condition, 18.44% of the RwD crashes resulted in fatalities whereas this figure is 11.46% for crashes occurring at dark with streetlight and 12.59% for the daylight condition. Young drivers of 15 to 24 years age group were more involved in crashes at dark no streetlight (34.68%) compared to the other two lighting conditions (daylight = 28.51%, dark with streetlight = 31.58%). Drivers have a tendency to not using seatbelts in dark no streetlight conditions (41.31%) compared to daylight (28.28%) and dark with streetlight (30.65%) while involving in a RwD crash. Some of the other variable categories are overrepresented across all three lighting conditions including moderate injury severity (81.02%, 76.47%, 74.25%), single-vehicle collision (90.11%, 89.16%, 95.29%), dry surface condition (84.45%, 82.97%, 84.97%), clear weather condition (72.68%, 74.30%, 74.96%), crashes occurring on weekday (67.93%, 55.11%, 57.19%), no abnormalities on the roadway (92.34%, 92.26%, 89.84%), straight roadways (56.40%, 57.59%, 57.31%), level vertical alignment (80.19%, 84.52%, 83.88%), speed limit between 40 and 55 mph (94.08%, 79.88%, 96.29%), male drivers (65.13%, 77.09%, 74.16%), driver protection system properly used (65.10%, 52.32%, 50.62%), no passenger presence (71.57%, 68.11%, 71.33%), and involving car/van/SUV (50.43%, 52.32%, 54.18%) type of vehicle. Note that, the percentages in parenthesis indicate the sequence (daylight, dark with streetlight, dark no streetlight).

*3.5 Variable Selection by Random Forest*

Variable selection is an important step to detect important variables to be fed into the Association Rules Mining algorithm. If there are too many variables, the model may identify unimportant crash patterns and learn from noise. In this study, the random forest (RF) algorithm was employed to discover significant variables from the dataset with a high importance value using Mean Decrease Accuracy (MDA) as a system of measurement (Han, Guo, & Yu, 2016). A variable importance plot suggests which variables had the highest effect in the classification model. In this case, 'lighting condition' was considered as a response variable and all other variables (total = 18) were considered explanatory variables. The more the accuracy of the random forest reduces as a consequence of removing (or permuting) a single variable, the more valuable that variable is assessed (Hur, Ihm, & Park, 2017). Note that, variables with a large MDA are more vital for the classification of the data. The following **Figure 3** shows the variable importance plot where each variable is shown on the y-axis and associated MDA on the x-axis. It is worth noting that, variables on the y-axis are ordered from most to least important (top to bottom).

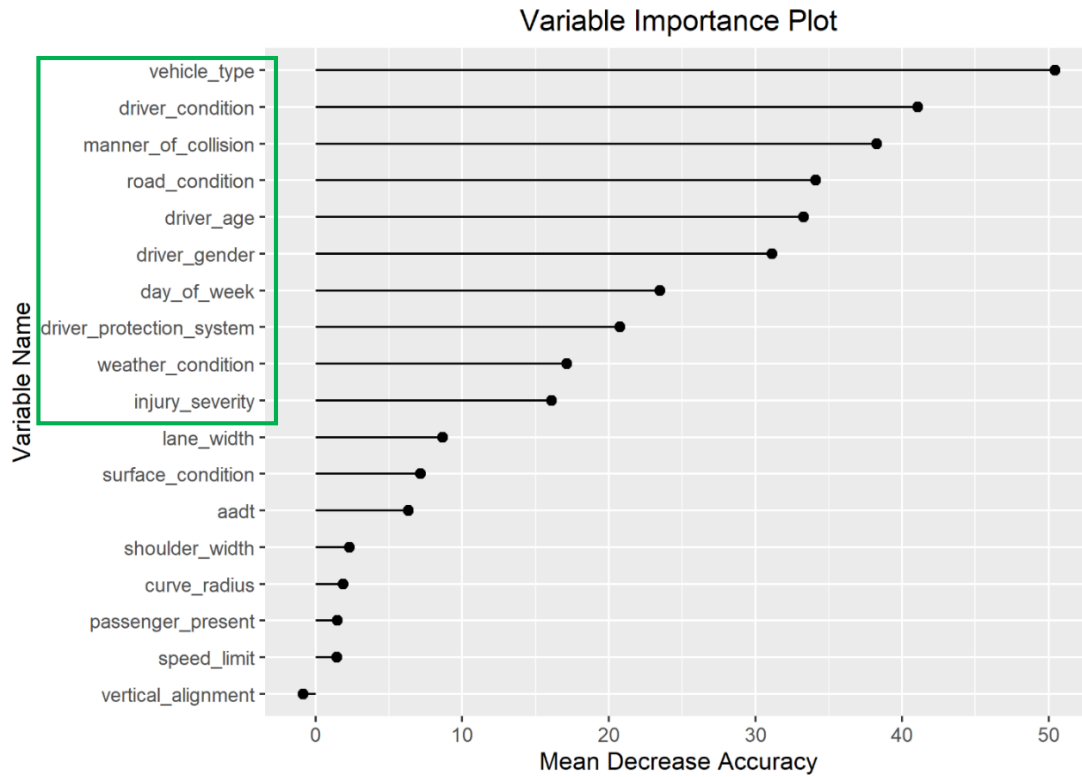

Figure 3. Variable importance plot by random forest algorithm

The top ten variables selected for further analysis were: vehicle type, driver condition, manner of collision, road condition, driver age, driver gender, day of week, driver protection system, weather condition, and injury severity.

## 4. Results and Discussion

According to the primary analysis, the final dataset consisted of 7,568 rows with 53 items (i.e., variable categories). The following relative item frequency plot (**Figure 4**) shows the relative frequencies of each item in the entire dataset. For example, the top item 'manner of collision = single vehicle' appeared 6,992 times in the entire database. Therefore, the relative frequency of this item is 6,992 divided by 7,568 which is 0.92. The other most frequently occurring items in the dataset were: road condition = no abnormalities (0.91), injury severity = moderate (0.77), weather condition = clear (0.73), driver gender = male (0.69) and so on.

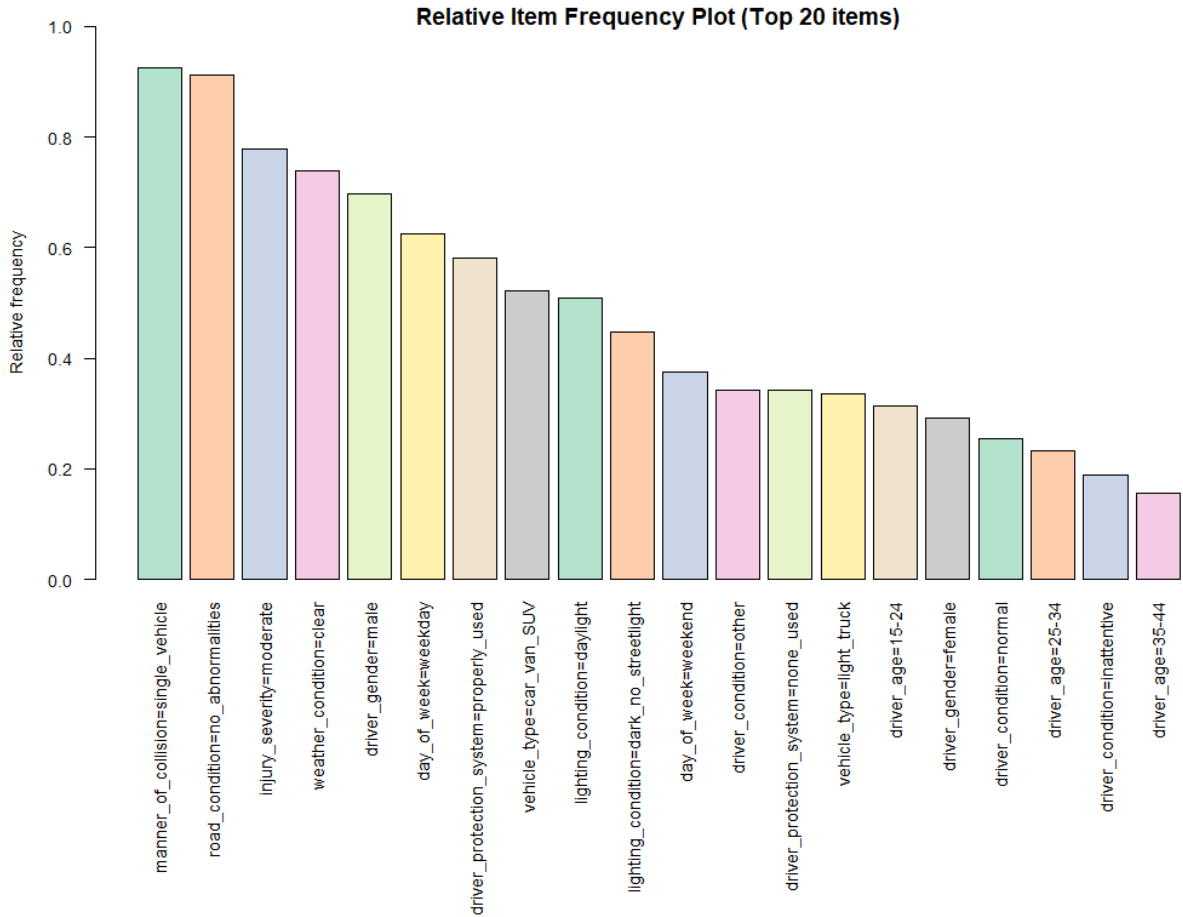

Figure 4. Relative item frequency plot

To generate meaningful rules using ARM, it is important to define an appropriate minimum threshold of the support and confidence parameters; otherwise, the algorithm could generate abundant decision rules. For example, without providing a threshold for support, confidence, and lift parameter, the algorithm generated a total of 1,724 rules. Graphical representations of these rules are provided in the following **Figure 5.**

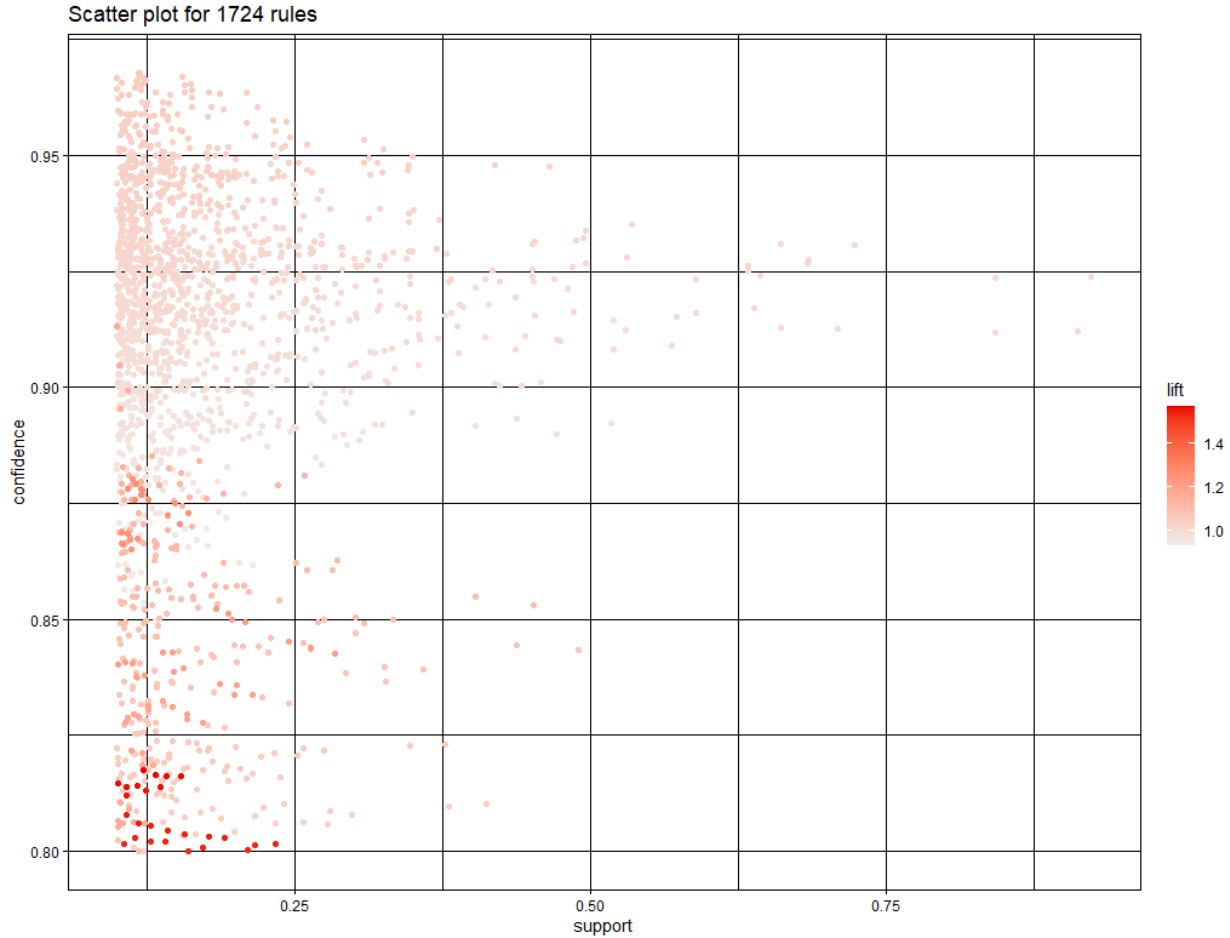

Figure 5. Graphical representation of association rules in terms of support, confidence, and lift

Researchers may require subject matter expertise to choose the lowest threshold values for support and confidence. After a substantial number of trials and errors, the minimum value of support and confidence was selected for each case. It might be claimed that the values of these parameters (support and confidence) were subjective and determined on a case-by-case basis (Das et al., 2019). A high lift value suggests a stronger relationship between the antecedents and consequents (Hornik, Grün, & Hahsler, 2005). Considering this point of view, a value of 1.1 was chosen as the minimum threshold for lift value. The investigation was restricted to 4-itemsets rules for ease of interpretation. A rule 'ID' was designed to identify and explain any given pattern associated with the generated rules. Using the selected variables, ARM was applied to three separate lighting condition scenarios.

*Case 1: Lighting Condition = Daylight*

The 'lighting_condition' variable was set to 'daylight' as the right-hand-side (RHS) to mine the association rules for case 1. After several trials and errors, the minimum value of support and confidence was set at 0.001% and 60% respectively. Initially, the algorithm generated 4,361 rules which contained a large number of redundant rules. To explore intuitive crash patterns, it is important to trim repetitive rules. After pruning redundant rules (Thabtah, 2005), 299 rules remained and were sorted according to descending order of lift value. **Table 4** lists the top 20 rules for this case.

Table 4. Top 20 Association rules for RwD crashes in daylight condition

| ID | Antecedent (s) | S (%) | C (%) | L |
|---|---|---|---|---|
| R1 | {injury_severity=fatal, weather_condition=cloudy, driver_condition=distracted} | 0.026 | 100.00 | 1.97 |
| R2 | {injury_severity=fatal, driver_age=45-54, road_condition=standing_water} | 0.013 | 100.00 | 1.97 |
| R3 | {injury_severity=fatal, weather_condition=cloudy, road_condition=construction} | 0.013 | 100.00 | 1.97 |
| R4 | {injury_severity=fatal, driver_age=35-44, road_condition=construction} | 0.013 | 100.00 | 1.97 |
| R5 | {driver_protection_system=none_used, road_condition=standing_water, driver_condition=distracted} | 0.013 | 100.00 | 1.97 |
| R6 | {driver_age=45-54, driver_protection_system=none_used, road_condition=standing_water} | 0.013 | 100.00 | 1.97 |
| R7 | {vehicle_type=truck} | 3.079 | 79.52 | 1.56 |
| R8 | {injury_severity=fatal, driver_protection_system=none_used, road_condition=construction} | 0.040 | 75.00 | 1.47 |
| R9 | {driver_age=35-44, driver_protection_system=properly_used, road_condition=construction} | 0.040 | 75.00 | 1.47 |
| R10 | {day_of_week=weekend, road_condition=standing_water, driver_condition=distracted} | 0.040 | 75.00 | 1.47 |
| R11 | {manner_of_collision=head_on} | 1.414 | 74.31 | 1.46 |
| R12 | {driver_age=>64} | 3.726 | 73.25 | 1.44 |
| R13 | {driver_gender=female, driver_protection_system=none_used, road_condition=standing_water} | 0.066 | 71.43 | 1.40 |
| R14 | {driver_age=45-54, road_condition=construction} | 0.092 | 70.00 | 1.38 |
| R15 | {road_condition=shoulder_abnormality} | 0.396 | 68.18 | 1.34 |
| R16 | {manner_of_collision=single_vehicle, weather_condition=cloudy, road_condition=construction} | 0.026 | 66.67 | 1.31 |
| R17 | {day_of_week=weekend, injury_severity=moderate, road_condition=construction} | 0.053 | 66.67 | 1.31 |
| R18 | {injury_severity=moderate, weather_condition=clear, road_condition=construction} | 0.211 | 64.00 | 1.26 |
| R19 | {driver_age=15-24, vehicle_type=car_van_SUV, driver_condition=distracted} | 0.423 | 64.00 | 1.26 |
| R20 | {day_of_week=weekday, weather_condition=rain, road_condition=standing_water} | 0.040 | 60.00 | 1.18 |

The first rule identified 'driver distraction' as a contributing factor resulting in fatalities in cloudy weather conditions during daylight. Three other driver distraction-related crash patterns were also identified (R5, R10, and R19). One of the previous research conducted in Louisiana identified that drivers driving on R2L highways were more prone to distracted driving (X. Sun, 2018). Some of the possible sources of distraction activity that may divert a driver's awareness from the task of driving include texting or talking over the cell phone, talking to the passenger(s), adjusting the audio, or navigation system (Alshatti, 2018). Four of the association rules (R2, R6, R13, and R20) offer another intriguing RwD collision pattern in the daylight triggered by 'standing water' on the road. Louisiana has a humid subtropical climate with annual precipitation of 64 inches, which is twice as much as the national average (Das, Dutta, & Sun, 2020). Again, poor drainage during rainfall reduces skid resistance greatly due to standing water on the road, causing hydroplaning and making vehicle steering harder, increasing the likelihood of RwD accidents (Park & Yun, 2017).

Eight (R3, R4, R8, R9, R14, R16, R17, and R18) of the top twenty association rules identified driver's involvement in RwD crashes at rural construction zones during the daylight with three (R3, R4, and R8) of them resulted in fatalities. Previous research suggests that work zone collisions are more common throughout the day (Li & Bai, 2006). The complexity of construction zones also plays a vital role as drivers usually have less time to undertake evasive maneuvers to avoid colliding with other vehicles or surrounding objects, increasing the risk of an accident (Ghasemzadeh & Ahmed, 2019). Two more rural construction zone-related RwD crashes during daylight condition were identified, both involving male drivers (R3, R18). There was additional evidence of RwD crashes involving older drivers (>64 years) during the daylight (R12, $L = 1.44$). One previous study has found that older drivers had a higher crash risk during the day, which also reflects this group's avoidance of nighttime driving (Fildes, 1994; Stutts, Martell, Staplin, & TransAnalytics, 2009). Along with impaired medical and physical conditions, inattention, and inappropriate turning are all potential triggers to be involved in a RwD crash involving older drivers (Liu & Subramanian, 2009).

Truck involvement in RwD collisions was recognized by one of the association rules (R7, $L = 1.56$). Truck driving is difficult on rural roads because they are frequently congested and only have two lanes. Additionally, large dimensions with heavy weight and lower deceleration in response to braking are some of the other factors associated with truck-involved crashes (Mohamedshah, Paniati, & Hobeika, 1993). When a car exits the roadway and enters the shoulder area, shoulder characteristics play a critical part in determining whether the driver will be involved in a crash or not. This issue was identified in rule 15. One prior study found that shoulder abnormalities (i.e., pavement edge drop-offs, holes, or potholes on the shoulder) possess a substantial risk for drivers to force into a RwD crash (Rahman et al., 2021). Drivers were also involved in daytime RwD crashes in the case of a head-on collision (R11, $L = 1.46$). A roadway departure or lane departure crash is often the mechanism for head-on crashes (Nelson et al., 2011).

*Case 2: Lighting Condition = Dark*

Out of the total 3,717 RwD crashes that occurred at dark on R2L highways, only 8.7% (323 crashes) of them occurred in presence of streetlight, while the remaining 91.3% (3,394 crashes) of them occurred in absence of streetlight. To address these two different lighting conditions at night, a binary variable 'streetlight' with outcome yes/no was incorporated into the model to reveal the associated crash patterns.

*Condition 1: Streetlight = Yes*

To mine the association rules for condition 1, the 'streetlight' variable was set to 'yes' as a consequent. Due to the lower sample size (323 crashes), the minimum value of support was set at 0.00005% following multiple rounds of trial and error. The minimum value of confidence was set as 55%. The algorithm initially produced 103 rules, many of which were repetitive. After pruning repetitive rules, 63 rules remained, which were sorted by lift value in descending order. The top 20 rules for this case are listed in following **Table 5**.

Table 5. Top 20 association rules for RwD crashes at dark with streetlight

| ID | Antecedent | S (%) | C (%) | L |
|---|---|---|---|---|
| R1 | {manner_of_collision=right_angle, driver_condition=alcohol} | 0.013 | 100 | 23.43 |
| R2 | {manner_of_collision=rear_end, injury_severity=severe, vehicle_type=car_van_SUV} | 0.013 | 100 | 23.43 |
| R3 | {day_of_week=weekend, manner_of_collision=rear_end, driver_age=15-24} | 0.013 | 100 | 23.43 |
| R4 | {injury_severity=severe, weather_condition=rain, road_condition=shoulder_abnormality} | 0.013 | 100 | 23.43 |
| R5 | {weather_condition=cloudy, driver_protection_system=improperly_used, driver_condition=alcohol} | 0.013 | 100 | 23.43 |
| R6 | {driver_age=15-24, driver_protection_system=improperly_used, driver_condition=alcohol} | 0.013 | 100 | 23.43 |
| R7 | {day_of_week=weekend, weather_condition=cloudy, driver_protection_system=improperly_used} | 0.013 | 100 | 23.43 |
| R8 | {injury_severity=severe, driver_age=15-24, driver_condition=distracted} | 0.013 | 100 | 23.43 |
| R9 | {injury_severity=severe, driver_age=45-54, vehicle_type=truck} | 0.026 | 100 | 23.43 |
| R10 | {manner_of_collision=rear_end, weather_condition=cloudy} | 0.026 | 50 | 11.72 |
| R11 | {manner_of_collision=rear_end, injury_severity=severe, driver_age=35-44} | 0.013 | 50 | 11.72 |
| R12 | {weather_condition=snow_sleet_hail, driver_gender=female, driver_protection_system=none_used} | 0.013 | 50 | 11.72 |
| R13 | {day_of_week=weekend, driver_protection_system=improperly_used, driver_condition=alcohol} | 0.013 | 50 | 11.72 |
| R14 | {driver_age=55-64, driver_gender=female, driver_protection_system=improperly_used} | 0.013 | 50 | 11.72 |
| R15 | {driver_gender=female, driver_protection_system=improperly_used, driver_condition=inattentive} | 0.013 | 50 | 11.72 |
| R16 | {manner_of_collision=single_vehicle, road_condition=unknown, driver_condition=drug} | 0.013 | 50 | 11.72 |
| R17 | {injury_severity=severe, vehicle_type=light_truck, driver_condition=distracted} | 0.013 | 50 | 11.72 |
| R18 | {injury_severity=fatal, driver_protection_system=unknown, driver_condition=alcohol} | 0.013 | 50 | 11.72 |
| R19 | {day_of_week=weekend, driver_age=55-64, driver_protection_system=improperly_used} | 0.013 | 50 | 11.72 |
| R20 | {injury_severity=severe, driver_protection_system=properly_used, vehicle_type=truck} | 0.026 | 40 | 9.37 |

At dark with streetlight, most of the association rules (R1, R5, R6, R14, R17, and R19) identified alcohol/drug involvement in RwD crashes on R2L highways. These rules were expected as alcohol/drugs could blur vision and impair the driver's ability to evaluate the space, speed, and movement of other vehicles (Royce & Scratchley, 1996). Furthermore, alcohol and other sedatives may impair the driver's ability to process information and react to critical driving situations (i.e., slow reaction time). With this type of impairment at night, the driver might drift across the centerline, move from one lane to another lane or even run off the road and get involved in a RwD crash (Shyhalla, 2014). It is worth mentioning that drivers did not properly use their seatbelts in three (R5, R6, and R14) of the alcohol-related RwD crashes that occurred in dark with streetlight. Previous research has identified a strong relationship between drinking and not wearing a seatbelt when involved in a crash (Desapriya, Pike, & Babul, 2006; Shyhalla, 2014; Tison, Williams, Chaudhary, & Group, 2010).

Two of the association rules identified young drivers (15-24 years) involved in RwD crashes at dark with streetlight (R3, R8). Excessive risk-taking behaviors, as well as other crash contributing factors including drinking and driving, speeding, fatigue or sleepiness, and inexperience, are all common reasons for a young driver to leave a travel lane at dark (J. Zhang, Lindsay, Clarke, Robbins, & Mao, 2000). Some of the other driver age groups were involved in RwD crashes in similar lighting conditions including (a) 45-54 years aged drivers in severe injury crashes while driving trucks (R9, L = 23.43) and (b) 34-44 years aged drivers in rear-end collisions resulting in severe injuries (R11, L = 11.72) (c) Female drivers between the age of 55 and 64 years who did not wear their seatbelts properly during the crash (R14, L = 11.72) and (d) 55-64 years aged drivers involved in RwD crashes over the weekend and wearing their seatbelts incorrectly during the RwD crash event.

According to the rule R4 (L = 24.43), the combination of rainy weather conditions and shoulder irregularities contributed to RwD crashes on R2L highways at dark with streetlight. Some of the other adverse weather-related RwD crashes were also identified (cloudy = R7, R10, snow/sleet/hail = R12). The deterioration of lane-keeping skills may be exacerbated by inclement weather due to reduced vision and slippery road surfaces (Allen, Parseghian, & Stein, 1996; Young, Regan, & Hammer, 2007). A lethal combination of reduced vision at night and cloudy weather conditions could result in a RwD crash. Two of the association rules (R15, R17) identified specific driver conditions (e.g., distraction, inattentive) responsible for RwD crashes on R2L highways at dark with streetlight. This is consistent with the results from the prior studies (Hashemi & Archilla, 2016; Liu & Subramanian, 2009).

*Condition 2: Streetlight = No*

To mine the association rules for condition 2, the 'streetlight' variable was set to 'no' as a consequent. Following multiple rounds of trial and error, the minimal values of support (0.004%) and confidence (55%) were selected. The algorithm initially produced 1,502 rules, many of which were repetitive. After pruning, 198 rules remained, which were sorted by lift value in descending order. The top 20 rules for this case are listed in **Table 6**.

Table 6. Top 20 association rules for RwD crashes at dark no streetlight

| ID | Antecedent | S (%) | C (%) | L |
|---|---|---|---|---|
| R1 | {manner_of_collision=head_on, injury_severity=severe, driver_gender=female} | 0.053 | 100.00 | 2.23 |
| R2 | {weather_condition=rain, driver_gender=male, road_condition=animal} | 0.172 | 100.00 | 2.23 |
| R3 | {injury_severity=moderate, vehicle_type=light_truck, road_condition=animal} | 0.145 | 91.67 | 2.04 |
| R4 | {injury_severity=severe, driver_age=35-44, driver_condition=ill_fatigued_asleep} | 0.053 | 80.00 | 1.78 |
| R5 | {manner_of_collision=single_vehicle, driver_gender=female, road_condition=animal} | 0.106 | 80.00 | 1.78 |
| R6 | {manner_of_collision=single_vehicle, driver_age=15-24, road_condition=animal} | 0.092 | 77.78 | 1.73 |
| R7 | {road_condition=animal} | 0.449 | 73.91 | 1.65 |
| R8 | {driver_age=25-34, vehicle_type=light_truck, driver_condition=drug} | 0.106 | 72.73 | 1.62 |
| R9 | {weather_condition=rain, vehicle_type=others, driver_condition=inattentive} | 0.066 | 71.43 | 1.59 |
| R10 | {vehicle_type=car_van_SUV, road_condition=unknown, driver_condition=inattentive} | 0.066 | 71.43 | 1.59 |
| R11 | {driver_condition=alcohol} | 5.959 | 69.28 | 1.54 |
| R12 | {manner_of_collision=head_on, driver_age=35-44, driver_protection_system=none_used} | 0.053 | 66.67 | 1.49 |
| R13 | {injury_severity=severe, driver_gender=male, driver_condition=ill_fatigued_asleep} | 0.145 | 64.71 | 1.44 |
| R14 | {day_of_week=weekend, driver_protection_system=none_used} | 8.100 | 62.30 | 1.39 |
| R15 | {weather_condition=cloudy, driver_age=35-44, driver_protection_system=none_used} | 0.436 | 62.26 | 1.39 |
| R16 | {injury_severity=fatal, driver_age=15-24} | 2.616 | 61.11 | 1.36 |
| R17 | {injury_severity=fatal, driver_condition=distracted} | 0.159 | 60.00 | 1.34 |
| R18 | {driver_gender=female, driver_protection_system=none_used, driver_condition=drug} | 0.119 | 60.00 | 1.34 |
| R19 | {driver_age=15-24, vehicle_type=light_truck, driver_condition=ill_fatigued_asleep} | 0.555 | 58.33 | 1.30 |
| R20 | {injury_severity=fatal, driver_protection_system=none_used} | 5.087 | 58.07 | 1.29 |

The first rule (R1, L = 2.23) suggests that female drivers were involved in a head-on collision at dark with no streetlight resulting in severe injuries. Previous research conducted by Morgan and Mannering identified female drivers' involvement in RwD crashes resulting in severe injury crashes while driving in dark and unlit conditions (Morgan & Mannering, 2011). Wildlife populations along R2L highways and animal activity in the dark might lead to a RwD crash. This intriguing crash pattern was backed up by five of the top twenty association rules (R2, R3, R5, R6, and R7). In the dark no streetlight condition, the driver might get shocked by the unforeseen animal(s) darting across the roadway and eventually lose control of the vehicle, resulting in a RwD crash (Rahman et al., 2021).

The physical condition of the driver has been recognized as a significant contributor to RwD collisions in dark settings. For example, 'Illness/fatigued/asleep' was identified as a crash contributing factor in several association rules (R4, R13, and R19) among which two of them resulted in severe injuries (R4, R13). Driving in the dark over lengthy, monotonous sections of R2L roadways may cause 'Hypovigilance', a condition in which the driver's focus is impaired (Larue, Rakotonirainy, & Pettitt, 2011). 'Hypovigilance' refers to a driver's decreased alertness or attentiveness and occurs due to the absence of stimuli (persistent, predictable, or highly repetitive) in the surrounding roadway environment (Larue, Rakotonirainy, & Pettitt, 2010). This increased monotony causes decreased information processing and a quick degradation in a driver's capacity to respond to unexpected situations, resulting in RwD accidents on R2L roadways at dark with no streetlight. Other physical conditions of drivers were found to be responsible for RwD accidents under similar illumination circumstances (inattentive = R9, R10 and alcohol/drug involvement = R8, R11).

At dark no streetlight condition, drivers have a tendency of not using seat-belt in several RwD crash scenarios including (a) involvement of 35-44 years aged drivers in a head-on collision (R12, $L = 1.49$), (b) during weekends (R14, $L = 1.39$) (c) involvement of 35-44 years aged drivers in cloudy weather conditions (R15, $L = 1.39$) (d) drug-involved female drivers (R18, $L = 1.34$) and (e) fatal injury severity (R20, $L = 1.29$). Three fatal RwD crash scenarios at night without streetlight were identified including the involvement of young drivers aged 15 to 24 years (R16, $L = 1.36$), driver distraction (R17, $L = 1.34$), and not using seat-belt (R20, $L = 1.29$). Previous research has identified that seat belt usage among fatally injured occupants was lower at dark than during the day (Tison et al., 2010).

## 5. Conclusions

Based on ten years (2008–2017) of fatal and injury RwD crashes on Louisiana's R2L highways, this study investigated the hidden crash patterns under three distinct illumination situations (daylight, dark with streetlight, dark no streetlight) by performing ARM. Some of the results verify the common perceptions of RwD crashes while few other discoveries were quite startling.

### 5.1 Research Findings

Some of the major findings from this investigation are summarized below:

- RwD departure crashes that occurred in dark lighting conditions, around 91.3 percent (3,394 of 3,717) of them occurred in absence of streetlight, suggesting that poor illumination plays a substantial role in RwD crashes.
- Most of the daylight RwD crashes were associated with rural construction zone and 'standing water' on the roadway.
- Older drivers of age group 65 or higher were more likely to involve in RwD crashes during the daylight.

- Driver physical condition (i.e., illness, fatigue, asleep, inattentive, distraction) and alcohol/drug involvement were identified as significant contributors to RwD crash at dark (with and without streetlight).
- Colliding with the animal in a dark no streetlight condition was also identified as an important RwD crash pattern.
- A strong association between alcohol/drug usage and no seat belt usage was recognized in RwD crashes during the dark (with and without streetlight).
- Drivers tend to improperly or not use a seat belt at dark with and without streetlight while involved in RwD crashes. Some of these 'no/improperly seat belt usage' related crashes were associated with alcohol intoxication, the young driver of age 15 to 24 years, and female drivers.

*5.2 Suggested Countermeasures*

Reducing RwD crashes necessitates problem-specific interventions. For example, 'standing water' was identified as an important contributor to RwD crashes during daylight. To tackle this, safety messages like "If it's Flooded, Forget It" can be effective in informing drivers of the dangers of driving in flooded areas (Clarkin, Keller, Warhol, & Hixson, 2006). Maintaining uniformity in roadway shoulder construction may help to accommodate off-tracking vehicles on R2L highways (Downs Jr & Wallace, 1982). Strict enforcement of the texting ban with awareness campaigns is expected to prevent distracted driving (X. Sun, 2018). The use of innovative traffic control devices (e.g., speed display trailers, variable message signs, radar drones, retroreflective vehicle visibility improvement) may help to alert drivers and improve safety in rural construction zones (Fontaine, Carlson, & Hawkins, 2000).

The majority of RwD crashes on R2L roadways were attributed to colliding with animals in dark lighting conditions. Although the available countermeasures to address animal-vehicle crashes are limited, some steps have been recommended in previous research, such as driving behavior (general education, warning signs, infrared detection from vehicles), roadway improvement (roadway lighting, roadside clearing, reduced speed limit, shoulder widening), and methods to influence animal behavior (physical control, underpasses, and overpasses, at-grade crosswalk) (Hedlund, Curtis, Curtis, & Williams, 2004). The co-occurrence of head-on collision and severe injury severity were identified for RwD crashes, implying that avoiding head-on collisions is crucial for minimizing the severity of RwD crashes especially in dark lighting condition. Installation of center line rumble strips (CLRS) is a proven safety countermeasure that may help to maintain the intended driving lane during poor visibility at night and reduce this type of head-on crash (Persaud, Retting, & Lyon, 2004).

The research team recognized driver alcohol/drug involvement as a serious safety concern in connection to RwD crashes. According to the database used in this study, a total of 282 individuals (daylight = 52, dark = 230) under the age of 21 years old were involved in RwD crashes while drinking alcohol. As the Minimum Legal Drinking Age (MLDA) in the United States is 21 years (Carpenter & Dobkin, 2011), strict enforcement of this rule would undoubtedly improve safety on R2L highways, particularly at dark. The research also discovered an intriguing driving behavior by identifying the relationship between alcohol/drug consumption and 'no seatbelt usage' while involved in a RwD crash at dark. Seatbelt use checkpoints with voluntary breath testing at night can be helpful in encouraging seatbelt use while also reducing drinking and driving (Solomon, Chaffe, Preusser, & Group, 2009). Other significant association rules revealed in this study can be reviewed by the safety professionals, who should seek techniques to break the association (chain of critical events) on a case-by-case basis.

Except for a case-by-case basis, some other additional countermeasures can be implemented on high crash cluster locations. For example, guardrails can be installed on locations where RwD crashes are frequent (Jalayer & Huaguo Zhou PHD, 2016). Median barriers are also another proven safety countermeasure that is extremely effective in reducing cross-median crashes and associated injury severity (Coulter & Ksaibati, 2013). According to the database used in this study, around 91% of the RwD crashes occurred in absence of streetlights at night suggesting that lighting sources can play a vital role in reducing such crashes. For example, adaptive LED streetlights can be installed in rural sharp curved roadways (C.-C. Sun et al., 2017). If heavy rains are predicted based on historical data and weather forecasts, some of the rural roads might be temporarily closed to avoid crashes associated with 'standing water' on the roadway.

*5.3 Research Contribution and practical applications*

Traditional techniques to identify crash contributing factors in RwD crashes are inadequate to capture the complex nature of accident characteristics, which might vary depending on lighting circumstances. As a result, further research is required from a different perspective to discover the relationship between crash risk variables and specific illumination conditions. This study utilized a 'safe system approach' to reveal the internal link among RwD crash contributing factors on R2L highways. For example, the study outcomes exhibited an important link between drivers' behavioral characteristics while involved in RwD crashes at dark (with and without streetlight) including a tendency towards violating mandatory driving regulations such as all-time seatbelt use, improper use of a seatbelt, alcohol/drug intake, and specific physical conditions (illness, fatigued, asleep). The knowledge of identified driving behavior patterns associated with RwD collisions can lead to effectively targeted driver education programs to mitigate risky driving maneuvers. Also, prioritizing crash factors of key patterns can help develop suitable countermeasures for improving safety.

*5.4 Limitations*

There are certain limitations to this research. The study was limited to four-itemset rules with a total of twenty rules for each case. More number of rules would result in additional intriguing patterns. The inclusion of other important variables related to the RwD crashes such as the presence of pavement markings, CLRS, SRS, seasonal difference, and so forth may help to discover the crash pattern more precisely.

There are several additional limitations of this study. Some of the identified crash patterns are based on the geographic characteristics of the state of Louisiana. For example, crashes due to 'standing water' on the roadway, may apply to other states which receive frequent precipitation (e.g., Hawaii, Mississippi, Alabama, Florida, Tennessee), and do not apply to states that receive the least amount of rain (e.g., Nevada, Arizona, Utah, New Mexico, Wyoming). Similarly, crashes associated with 'hitting animals' may apply to states that are top for animal-vs-vehicle collisions (e.g., West Virginia, Montana, South Dakota, Michigan, and Pennsylvania).

**Acknowledgments**

The authors would like to acknowledge the support of the Louisiana Department of Transportation and Development for supplying the database utilized in this study.

# References

Agrawal, R., Imieliński, T., & Swami, A. (1993). Mining association rules between sets of items in large databases. *Proceedings of the 1993 ACM SIGMOD International Conference on Management of Data*, 207–216.

Al-Bdairi, N. S. S., & Hernandez, S. (2017). An empirical analysis of run-off-road injury severity crashes involving large trucks. *Accident Analysis & Prevention*, *102*, 93–100.

Al-Bdairi, N. S. S., Hernandez, S., & Anderson, J. (2018). Contributing factors to run-off-road crashes involving large trucks under lighted and dark conditions. *Journal of Transportation Engineering, Part A: Systems*, *144*(1), 4017066.

Allen, R. W., Parseghian, Z., & Stein, A. C. (1996). A driving simulator study of the performance effects of low blood alcohol concentration. *Proceedings of the Human Factors and Ergonomics Society Annual Meeting*, *40*(18), 943–946. SAGE Publications Sage CA: Los Angeles, CA.

Alshatti, D. A. (2018). *Examining Driver Risk Factors in Road Departure Conflicts Using SHRP2 Data*. University of Dayton.

Appiah, J., & Zhao, M. (2020). *Examination of Features Correlated With Roadway Departure Crashes on Rural Roads*. Virginia. Dept. of Transportation.

Carpenter, C., & Dobkin, C. (2011). The minimum legal drinking age and public health. *Journal of Economic Perspectives*, *25*(2), 133–156.

Clarkin, K., Keller, G., Warhol, T., & Hixson, S. (2006). *Low-water crossings: geomorphic, biological, and engineering design considerations*.

Coulter, Z., & Ksaibati, K. (2013). *Effectiveness of various safety improvements in reducing crashes on Wyoming roadways*. Mountain Plains Consortium.

Council, F. M., Reurings, M., Srinivasan, R., Masten, S., & Carter, D. (2010). *Development of a Speeding-Related Crash Typology*. Turner-Fairbank Highway Research Center.

Das, S., Dutta, A., Avelar, R., Dixon, K., Sun, X., & Jalayer, M. (2019). Supervised association rules mining on pedestrian crashes in urban areas: identifying patterns for appropriate countermeasures. *International Journal of Urban Sciences*, *23*(1), 30–48.

Das, S., Dutta, A., & Sun, X. (2020). Patterns of rainy weather crashes: Applying rules mining. *Journal of Transportation Safety & Security*, *12*(9), 1083–1105.

Desapriya, E., Pike, I., & Babul, S. (2006). Public attitudes, epidemiology and consequences of drinking and driving in British Columbia. *IATSS Research*, *30*(1), 101–110.

Downs Jr, H. G., & Wallace, D. W. (1982). *Shoulder geometrics and use guidelines*.

Eustace, D., Almutairi, O. E., & Hovey, P. W. (2016). Modeling factors contributing to injury and fatality of run-off-road crashes in Ohio. *Advances in Transportation Studies, Section B*, *40*.

Fildes, B. (1994). *Older Road User Crashes Report No. 61*.

Fontaine, M. D., Carlson, P. J., & Hawkins, H. G. (2000). *Use of Innovative Traffic Control Devices to Improve Safety at Short-Term Rural Work Zones*. Texas Transportation Institute, Texas A & M University System.

Ghasemzadeh, A., & Ahmed, M. M. (2019). Exploring factors contributing to injury severity at work zones considering adverse weather conditions. *IATSS Research*, *43*(3), 131–138.

Gibbons, R. B., Milton, J. C., & van Schalkwyk, I. (2016). *Impact of Roadway Lighting on Nighttime Crash Performance and Driver Behavior*.

Hahsler, M. (2017). arulesViz: Interactive Visualization of Association Rules with R. *R J.*, *9*(2), 163.

Hahsler, M., Chelluboina, S., Hornik, K., & Buchta, C. (2011). The arules R-package ecosystem: analyzing interesting patterns from large transaction data sets. *The Journal of Machine Learning Research*, *12*, 2021–2025.

Han, H., Guo, X., & Yu, H. (2016). Variable selection using mean decrease accuracy and mean decrease gini based on random forest. *2016 7th Ieee International Conference on Software Engineering and Service Science (Icsess)*, 219–224. IEEE.

Hashemi, M., & Archilla, A. R. (2016). Potential Factors Affecting Roadway Departure Crashes in Oahu, Hawaii. *ITE Western District Annual Meeting, Albuquerque*, 10.

Hashemi, M., & Archilla, A. R. (2017). Exploratory Analysis of Roadway Departure Crashes Contributing Factors Based on Classification and Regression Trees. *ITE Western District Annual Meeting, San Diego*, *55*(June), 1–8.

Hedlund, J. H., Curtis, P. D., Curtis, G., & Williams, A. F. (2004). Methods to reduce traffic crashes involving deer: what works and what does not. *Traffic Injury Prevention*, *5*(2), 122–131.

Hingson, R., & Winter, M. (2003). Epidemiology and consequences of drinking and driving. *Alcohol Research & Health*, *27*(1), 63.

Hong, J., Tamakloe, R., & Park, D. (2020). Discovering insightful rules among truck crash characteristics using apriori algorithm. *Journal of Advanced Transportation*, *2020*.

Hornik, K., Grün, B., & Hahsler, M. (2005). arules-A computational environment for mining association rules and frequent item sets. *Journal of Statistical Software*, *14*(15), 1–25.

Hossain, A., Sun, X., Thapa, R., & Codjoe, J. (2022). Applying Association Rules Mining to Investigate Pedestrian Fatal and Injury Crash Patterns Under Different Lighting Conditions: *Https://Doi.Org/10.1177/03611981221076120*, 036119812210761. https://doi.org/10.1177/03611981221076120

Hossain, M. M., Rahman, M. A., Sun, X., & Mitran, E. (2021). Investigating underage alcohol-intoxicated driver crash patterns in Louisiana. *Transportation Research Record*, *2675*(11), 769–782.

Hossain, M. M., Sun, X., Mitran, E., & Rahman, M. A. (2021). Investigating fatal and injury crash patterns of teen drivers with unsupervised learning algorithms. *IATSS Research*. https://doi.org/10.1016/J.IATSSR.2021.07.002

Hossain, M. M., Zhou, H., Das, S., Sun, X., & Hossain, A. (2022). Young drivers and cellphone distraction: Pattern recognition from fatal crashes. *Journal of Transportation Safety & Security*, 1–26. https://doi.org/10.1080/19439962.2022.2048763

Hsu, Y.-T., & Chang, S.-C. (2020). Using Association Rule Mining to Analyze the Accident Characteristics of Intersection with Different Control Types. *International Journal of Applied Science and Technology*, *10*(1), 45–54. https://doi.org/10.30845/ijast.v10n1p6

Hur, J.-H., Ihm, S.-Y., & Park, Y.-H. (2017). A variable impacts measurement in random forest for mobile cloud computing. *Wireless Communications and Mobile Computing*, *2017*.

Islam, M., & Pande, A. (2020). Analysis of single-vehicle roadway departure crashes on rural curved segments accounting for unobserved heterogeneity. *Transportation Research Record*, *2674*(10), 146–157.

Jägerbrand, A. K., & Sjöbergh, J. (2016). Effects of weather conditions, light conditions, and road lighting on vehicle speed. *SpringerPlus*, *5*(1), 1–17.

Jalayer, M., & Huaguo Zhou PHD, P. E. (2016). Overview of safety countermeasures for roadway departure crashes. *Institute of Transportation Engineers. ITE Journal*, *86*(2), 39.

Kim, J.-K., Ulfarsson, G. F., Kim, S., & Shankar, V. N. (2013). Driver-injury severity in single-vehicle crashes in California: a mixed logit analysis of heterogeneity due to age and gender. *Accident Analysis & Prevention*, *50*, 1073–1081.

Kotsiantis, S., & Kanellopoulos, D. (2006). Association rules mining: A recent overview. *GESTS International Transactions on Computer Science and Engineering*, *32*(1), 71–82.

Kusano, K. D., & Gabler, H. C. (2013). Characterization of opposite-direction road departure crashes in the united states. *Transportation Research Record*, *2377*(1), 14–20.

LADOTD. (2018). *Destination Zero Deaths: Louisiana Strategic Highway Safety Plan*.

Retrieved from http://www.destinationzerodeaths.com/Images/Site Images/ActionPlans/SHSP.pdf

Langford, J., & Oxley, J. (2006). Using the safe system approach to keep older drivers safely mobile. *IATSS Research*, *30*(2), 97–109.

Larue, G. S., Rakotonirainy, A., & Pettitt, A. N. (2010). Predicting driver's hypovigilance on monotonous roads: literature review. *1st International Conference on Driver Distraction and Inattention*.

Larue, G. S., Rakotonirainy, A., & Pettitt, A. N. (2011). Driving performance impairments due to hypovigilance on monotonous roads. *Accident Analysis & Prevention*, *43*(6), 2037–2046. https://doi.org/10.1016/J.AAP.2011.05.023

Li, Y., & Bai, Y. (2006). Fatal and injury crash characteristics in highway work zones. *Proc., Transportation Research Board 87 Th Annual Meeting*.

Liu, C., & Subramanian, R. (2009). *Factors related to fatal single-vehicle run-off-road crashes*.

Lord, D., Brewer, M. A., Fitzpatrick, K., Geedipally, S. R., & Peng, Y. (2011). *Analysis of roadway departure crashes on two lane rural roads in Texas.* Texas Transportation Institute.

Massie, D. L. (1993). *Analysis of accident rates by age, gender, and time of day based on the 1990 Nationwide Personal Transportation Survey. Final report*.

McGee Sr, H. W. (2018). *Practices for preventing roadway departures*.

Mohamedshah, Y. M., Paniati, J. F., & Hobeika, A. G. (1993). Truck accident models for interstates and two-lane rural roads. *Transportation Research Record*, *1407*, 35–41.

Morgan, A., & Mannering, F. L. (2011). The effects of road-surface conditions, age, and gender on driver-injury severities. *Accident Analysis & Prevention*, *43*(5), 1852–1863.

National Highway Traffic Safety Administration. (2002). *Traffic Safety Facts 2002: Alcohol*.

National Safety Council. (2010). *KABCO Injury Classification Scale and Definitions KABCO Injury Classification Scale and Definitions*.

Nelson, M., Miller, J. P., Zisman, I., Isackson, C., Helms, D., Albin, R. B., & Focke, D. (2011). *Best Practices In Lane-Departure Avoidance and Traffic Calming*.

Neuner, M., Atkinson, J. E., Chandler, B. E., Hallmark, S. L., Milstead, R., & Retting, R. (2016). *Integrating Speed Management within Roadway Departure, Intersections, and Pedestrian and Bicyclist Safety Focus Areas*. United States. Federal Highway Administration. Office of Safety.

Pahukula, J., Hernandez, S., & Unnikrishnan, A. (2015). A time of day analysis of crashes involving large trucks in urban areas. *Accident Analysis & Prevention*, *75*, 155–163.

Park, J., & Yun, D. (2017). *Analysis of road accident according to road surface condition*.

Peng, Y., Geedipally, S. R., & Lord, D. (2012). Effect of roadside features on single-vehicle roadway departure crashes on rural two-lane roads. *Transportation Research Record*, *2309*(1), 21–29.

Persaud, B. N., Retting, R. A., & Lyon, C. A. (2004). Crash reduction following installation of centerline rumble strips on rural two-lane roads. *Accident Analysis & Prevention*, *36*(6), 1073–1079.

Plainis, S., Murray, I. J., & Pallikaris, I. G. (2006). Road traffic casualties: understanding the night-time death toll. *Injury Prevention*, *12*(2), 125–138.

Rahman, M. A., Sun, X., Das, S., & Khanal, S. (2021). Exploring the influential factors of roadway departure crashes on rural two-lane highways with logit model and association rules mining. *International Journal of Transportation Science and Technology*, *10*(2), 167–183. https://doi.org/10.1016/J.IJTST.2020.12.003

Royce, J. E., & Scratchley, D. (1996). *Alcoholism and other drug problems*. Simon and Schuster.

Shyhalla, K. (2014). Alcohol involvement and other risky driver behaviors: effects on crash initiation and crash severity. *Traffic Injury Prevention*, *15*(4), 325–334.

Solomon, M. G., Chaffe, R. H. B., Preusser, D. F., & Group, P. R. (2009). *Nighttime enforcement of seat belt laws: An evaluation of three community programs*. United States. National Highway Traffic Safety Administration.

Spainhour, L. K., & Mishra, A. (2008). Analysis of fatal run-off-the-road crashes involving overcorrection. *Transportation Research Record*, *2069*(1), 1–8.

Stutts, J., Martell, C., Staplin, L., & TransAnalytics, L. L. C. (2009). *Identifying behaviors and situations associated with increased crash risk for older drivers*.

Sun, C.-C., Lee, X.-H., Moreno, I., Lee, C.-H., Yu, Y.-W., Yang, T.-H., & Chung, T.-Y. (2017). Design of LED street lighting adapted for free-form roads. *IEEE Photonics Journal*, *9*(1), 1–13.

Sun, X. (2018). *Investigating Problem of Distracted Drivers on Louisiana Roadways*.

Sun, X., & Rahman, M. A. (2021). *Impact of Center Line Rumble Strips and Shoulder Rumble Strips on all Roadway Departure Crashes in Louisiana Two-lane Highways*.

Terry, T. N., Brimley, B. K., Gibbons, R. B., & Carlson, P. J. (2016). *Roadway visibility research needs assessment*. United States. Federal Highway Administration. Office of Safety Research and ….

Thabtah, F. (2005). Rules pruning in associative classification mining. *Proceedings of the IBIMA Conference*, 7–15. Citeseer.

Tison, J., Williams, A. F., Chaudhary, N. K., & Group, P. R. (2010). *Daytime and nighttime seat belt use by fatally injured passenger vehicle occupants.* United States. National Highway Traffic Safety Administration.

Turochy, R. E., & Ozelim, L. (2016). *A study of the effects of pavement widening, rumble strips, and rumble stripes on rural highways in Alabama*. Auburn University. Highway Research Center.

Ünvan, Y. A. (2021). Market basket analysis with association rules. *Communications in Statistics-Theory and Methods*, *50*(7), 1615–1628.

Wang, J.-S., & Knipling, R. R. (1994). *Single-vehicle roadway departure crashes: problem size assessment and statistical description*. National Highway Traffic Safety Administration.

Weng, J., Zhu, J.-Z., Yan, X., & Liu, Z. (2016). Investigation of work zone crash casualty patterns using association rules. *Accident Analysis & Prevention*, *92*, 43–52.

Young, K., Regan, M., & Hammer, M. (2007). Driver distraction: A review of the literature. *Distracted Driving*, *2007*, 379–405.

Yu, M., Ma, C., & Shen, J. (2021). Temporal stability of driver injury severity in single-vehicle roadway departure crashes: a random thresholds random parameters hierarchical ordered probit approach. *Analytic Methods in Accident Research*, *29*, 100144.

Yu, S., Jia, Y., & Sun, D. (2019). Identifying Factors that Influence the Patterns of Road Crashes Using Association Rules: A case Study from Wisconsin, United States. *Sustainability 2019, Vol. 11, Page 1925*, *11*(7), 1925. https://doi.org/10.3390/SU11071925

Zhang, J., Lindsay, J., Clarke, K., Robbins, G., & Mao, Y. (2000). Factors affecting the severity of motor vehicle traffic crashes involving elderly drivers in Ontario. *Accident Analysis & Prevention*, *32*(1), 117–125.

Zhang, K., & Hassan, M. (2019). Crash severity analysis of nighttime and daytime highway work zone crashes. *PLoS One*, *14*(8), e0221128.

Zhu, H., Dixon, K. K., Washington, S., & Jared, D. M. (2010). Predicting single-vehicle fatal crashes for two-lane rural highways in Southeastern United States. *Transportation Research Record*, *2147*(1), 88–96.